\pdfoutput=1

\documentclass[11pt]{article}

\makeatletter
{\small
\xdef\f@size@small{\f@size}
\xdef\f@baselineskip@small{\f@baselineskip}
\normalsize
\xdef\f@size@normalsize{\f@size}
\xdef\f@baselineskip@normalsize{\f@baselineskip}
}

\newcommand{\semismall}{
  \fontsize
    {\fpeval{(\f@size@small+\f@size@normalsize)/2}}
    {\fpeval{(\f@baselineskip@small+\f@baselineskip@normalsize)/2}}
  \selectfont
}
\makeatother

\usepackage[preprint]{acl}

\usepackage{times}
\usepackage{latexsym}
\usepackage{booktabs}
\usepackage{array}
\usepackage{amssymb}
\usepackage{amsmath}
\usepackage{hyperref}
\usepackage{subcaption}
\usepackage{comment}
\usepackage{enumitem}
\usepackage{listings}
\usepackage[T1]{fontenc}
\usepackage[utf8]{inputenc}
\usepackage{microtype}
\usepackage{inconsolata}
\usepackage{graphicx}

\definecolor{beige}{rgb}{0.96, 0.96, 0.86}
\definecolor{green}{rgb}{0.07, 0.53, 0.03}
\definecolor{red}{rgb}{0.8, 0.25, 0.33}
\definecolor{NavyBlue}{HTML}{006EB8}
\lstdefinelanguage{Python}{
identifierstyle=\color{black},
comment=[l]{\#}, commentstyle=\color{gray}\ttfamily,
basicstyle=\semismall\ttfamily,
breaklines=True, breakatwhitespace=True,
upquote=True,
frame=single,
backgroundcolor=\color{beige},
columns=fullflexible, linewidth=0.975\columnwidth
}

\title{\emph{Not (yet) the whole story}: Evaluating Visual Storytelling Requires More than Measuring Coherence, Grounding, and Repetition}

\author{Aditya K Surikuchi, Raquel Fern{\'a}ndez, Sandro Pezzelle\\
        Institute for Logic, Language and Computation\\
        University of Amsterdam\\
        \texttt{\{a.k.surikuchi|raquel.fernandez|s.pezzelle\}@uva.nl}}

\begin{document}
\maketitle
\begin{abstract}

Visual storytelling consists in generating a natural language story given a temporally ordered sequence of images. This task is not only challenging for models, but also very difficult to evaluate with automatic metrics since there is no consensus about what makes a story `good'. In this paper, we introduce a novel method that measures story quality in terms of human likeness regarding three key aspects highlighted in previous work: visual grounding, coherence, and repetitiveness. We then use this method to evaluate the stories generated by several models, showing that the foundation model LLaVA obtains the best result, but only slightly so compared to TAPM, a 50-times smaller visual storytelling model. Upgrading the visual and language components of TAPM results in a model that yields competitive performance with a relatively low number of parameters. Finally, we carry out a human evaluation study, whose results suggest that a `good' story may require more than a human-like level of visual grounding, coherence, and repetition.

\end{abstract}

\section{Introduction}
\label{sec:1}

Visual storytelling is the task of generating a story for a sequence of several temporally-ordered images or video frames. For both human speakers and machine learning models, the task requires connecting the visual data causally, to generate a narrative consistent with the contents of the images. As for model-generated stories, evaluation is one of the key challenges due to the inherently creative nature of the task. Since human-written stories are typically used to train visual storytelling models---under the assumption that these stories provide a good learning signal---most previous work evaluated model-generated stories by directly comparing them to human ones using pattern-matching metrics. However, this approach is simplistic as it ignores several key aspects of visual stories, such as their degree of visual grounding, their overall coherence, or how repetitive they are. This problem has only been addressed recently, with \citet{RoViST} and \citet{GROOViST} proposing various metrics to take into account some of these crucial aspects. These methods assess the appropriateness of a generated story independently from its overlap with a ground-truth story for the same image sequence. Given that the same image sequence can possibly give rise to many different stories, this type of higher-level evaluation that does not rely on text overlap is clearly desirable.

Nevertheless, we argue that measuring the degree of coherence or visual grounding of a story may not be sufficiently informative, as there are no standard conventions that determine the preferable level of such properties. To address this issue, in this work, we first propose an evaluation method that assesses the quality of generated stories in terms of their \textit{distance} from human-written stories along several relevant dimensions, each measured by an available reference-free metric. Using this method, we evaluate a range of models on the visual storytelling task, including models specifically designed and trained for this task, as well as---for the first time---foundation models pre-trained to achieve general-purpose language and vision abilities, which we test in a zero-shot manner. We show that LLaVA \citep{llava-v1.6}, a powerful foundation model, performs best on the task, but only slightly better than TAPM \citep{TAPM}, a model designed for visual storytelling which is 50 times smaller than LLaVA. Second, given insights derived from our proposed distance-based evaluation method, we upgrade the visual and language components on TAPM, resulting in a model that achieves comparable performance to LLaVA with a significantly lower number of parameters. 

Our results show that the stories generated by LLaVA and the upgraded TAPM model are very close to human stories regarding their degree of visual grounding, coherence, and repetition. To further make sense of this finding, we collect human judgments with regards to comparing human and model stories. The results of this qualitative study indicate that humans tend to prefer human-written stories by a significant margin despite the quantitative closeness we observe.

In sum, we make the following contributions:

\begin{itemize}[noitemsep,topsep=0pt]
    \item We propose a novel evaluation method for visual storytelling quantifying the distance between human-written and model-generated stories in terms of visual grounding, coherence, and non-repetitiveness.
    \item We use our evaluation to assess the stories generated by various visual storytelling-specific and, for the first time in the community, general-purpose foundation models; we report the novel finding that a foundation model, LLaVA, achieves the best result (lowest distance) when prompted under a novel setting.
    \item We leverage insights from this finding to upgrade a visual storytelling-specific model, TAPM, by replacing its visual and language components; we show that doing so results in better performance, on par with or outperforming the best-performing (and twice larger) LLaVA model.
    \item Through human evaluation, we validate the scores of our distance-based method; at the same time, we report that human-written stories are still preferred, which suggests that the ingredients for a good story may not be limited to a human-like level of visual grounding, coherence, and non-repetitiveness.
\end{itemize}

Our code is available at: \url{https://github.com/akskuchi/dHM-visual-storytelling}.

\section{Related Work}
\label{sec:2}

\subsection{Visual Storytelling}
\label{sec:2_1}

Computational work on visual storytelling was initiated by \citet{vist-dataset}, who operationalized the task as generating a textual story given an ordered sequence of images. The authors proposed the VIST dataset, which comprises sequences of five natural images collected from Flickr albums, with corresponding stories provided by human crowd-workers. VIST has been a catalyst for developing visual storytelling models \cite{GLACNet, AREL, TAPM}. Over the last few years, other datasets have emerged that differ from VIST in some key features. On the one hand, to limit the complexity of modelling real-world knowledge implicit in natural images, \citet{AESOP} proposed AESOP, a dataset that includes sequences of three synthetic images (constructed by crowd-workers using clip–art entities from Abstract Scenes by~\citet{abstract_scenes}) and corresponding three long-paragraph stories. More recently, to overcome the possible lack of character consistency resulting from sampling images from Flickr albums, \citet{VWP} proposed the VWP dataset, which comprises sequences of movie shots including 5-10 images with corresponding stories provided by crowd-workers.

Regarding modeling, various computational approaches using RNNs and Transformers have been proposed for the task of generating plausible stories. Some of these models are trained end-to-end on the VIST dataset \cite{GLACNet, AREL, TAPM}, while other approaches utilize external knowledge sources \cite{KE-VIST, PR-VIST, MCSM-BART}. We describe some of these models in detail in Section~\ref{sec:4_2}. Regardless of the specific architectures, a challenge common to all computational approaches to this task is evaluation. In the following subsection, we review existing work on visual story evaluation in the general context of evaluating visually-grounded language.

\subsection{Visually-Grounded Language Evaluation}
\label{sec:2_2}

Since visual storytelling is essentially a vision-to-language task similar to video/image captioning, evaluation of generated stories typically employed reference-based pattern-matching metrics such as METEOR \cite{METEOR} and CIDEr \cite{CIDEr}. However, these $n$-gram based metrics are shown to correlate poorly with human judgments \cite{nlg-metrics-efficacy, AREL}. Other reference-based evaluation metrics such as BERTScore \cite{BERTScore} and BLEURT \cite{BLEURT} have also been used. These leverage pre-trained models to compute similarities between the generated candidate text and the corresponding references in a high-dimensional embedding space, which makes them more flexible to paraphrases and synonyms compared to standard $n$-gram based metrics. Nevertheless, reference-based metrics are by design only suitable for target-oriented generation tasks (e.g., machine translation), where it is considered essential for the generated text to match a curated set of references \cite{MAUVE,giulianelli-etal-2023-comes}. This is not the case in visual storytelling, where several stories could be plausible for a given image sequence. Recently, CLIPScore \cite{CLIPScore} has been proposed to quantify the degree of alignment between an image and a given text without the need for any reference. It computes a similarity score between the CLIP \cite{CLIP} embeddings of the image and the text, and has been shown to correlate well with human judgments. As such, it is widely adopted for evaluation in the image captioning task. However, its application in the domain of visual storytelling is less straightforward, as a visual sequence comprises multiple images, and stories are made up of multiple related sentences, which makes evaluating what counts as a  `good' story very challenging.

To address some of these challenges, \citet{RoViST} proposed a metric---RoViST---specifically for the visual storytelling task, which assesses three aspects of generated stories: visual grounding, coherence, and repetition. Subsequently, \citet{GROOViST} proposed GROOViST, a more advanced method to evaluate grounding in visual storytelling. However, while this constitutes important progress, in practice it is often not easy to make sense of these metrics, since they do not offer a reference point for assessing the quality of a generated story. For example, should stories be 100\% grounded in the images? Should they exhibit zero levels of repetition? We argue that applying the metrics to both \textit{model}- and \textit{human}-generated stories is essential to boost their capacity to evaluate story-generation models. In the next section, we propose a novel method that addresses this problem by building on the existing metrics proposed by \citet{RoViST} and \citet{GROOViST}.

\section{Problem Formulation}
\label{sec:3}

In this work, we take a human-centric approach and define the quality of generated stories in terms of  their `closeness' to stories produced by humans, regarding different dimensions that capture abstract properties of interest. Concretely, we compute dimension-specific scores for model- and for human-generated stories, measure human-model distance per dimension, and aggregate these distances to derive an overall distance score. We posit that the lower the overall distance, the more the generated story complies with high-level features observed in human stories.

Deciding on which dimensions are most relevant to determine the quality of a story, let alone how to operationalize such dimensions, is not trivial. In this study, we consider the three aspects proposed by \citet{RoViST}: visual grounding, coherence, and repetition. Here we describe how they are operationalized as reference-free metrics, following which we formally define our proposed method.

\subsection{Metrics}
\label{sec:metrics}

\paragraph{Visual Grounding}
To measure the degree of visual grounding of a story, we use GROOViST \citep{GROOViST}. For a given <image-sequence, story> pair, GROOViST first computes the alignment between the noun phrases (NPs) in the story and the bounding boxes in the images using their corresponding CLIP \cite{CLIP} embeddings. For each NP, only the maximum visual alignment score is retained. To penalize NPs with low visual alignment scores, the mean score of all the NPs in the dataset is used as a threshold.
Specifically, this step is implemented by calculating the distance of each NP’s score from the threshold. Resulting scores of the NPs are then weighted using word concreteness ratings to differentiate abstract words from concrete ones. The overall visual grounding score of a story is the sum of the concreteness weighted scores of all NPs normalized by the total number of NPs in the story. The resulting scores are bound to range $[-1, 1]$ with higher values indicating greater degree of visual grounding.

\paragraph{Coherence} 
We use a slightly modified version of RoViST-C \citep{RoViST} to evaluate the coherence of a given story, which corresponds to the average probability with which each sentence follows the preceding sentences. These probabilities are computed using ALBERT \cite{ALBERT} fine-tuned for the sentence order prediction task using the VIST and ROCStories \cite{ROCStories} datasets. For each sentence ($s_i$) in a story, we obtain the probability that it follows the entire concatenated prefix of all previous sentences ($\{s_{1},..., s_{i-1}\}$)---instead of just the previous sentence ($s_{i-1}$) as done in the original RoViST-C. The overall coherence score of a story is obtained by taking the average of these probabilities across all its sentences resulting in a value between 0 and 1 (indicates high coherence).

\paragraph{Repetition}
We measure the degree of repetition of a story using the RoViST-NR metric, where NR stands for `non-redundancy' \cite{RoViST}. For two segments of text, repetition is computed using the Jaccard Similarity (JS) \cite{JS_for_text} which is defined as the number of co-occurring words between the two texts normalized by the total number of words in both texts. Inter-sentence repetition is obtained as the average of JS scores computed between each sentence $s_i$ and all its preceding sentences ($\{s_{1},..., s_{i-1}\}$). For every sentence in the story, the intra-sentence repetition is obtained by computing the average of JS scores between non-overlapping $4$-gram phrases of the sentence. The overall repetition score of a story is the average of all inter- and intra-sentence scores subtracted from $1$. The resulting scores range between $0$ and $1$ and stories with scores closer to $1$ indicate less repetition.\looseness-1

\subsection{Distance between Humans and Models}
\label{sec:3_1}

For a given <image sequence, model story> pair, we compute the coherence $C_M$, visual grounding $G_M$, and repetition $R_M$ scores using the three metrics described above. We do the same for the corresponding <image sequence, human story> pair, and denote the resulting scores as $C_H$, $G_H$, and $R_H$. We then compute the absolute differences between the human stories and the model-generated ones to measure metric-level deviations:
\begin{equation}
\label{eq:3a}
\begin{split}
    \text{d}^{C}_{HM}&=|C_H - C_M|,\\
    \text{d}^{G}_{HM}&=|G_H - G_M|,\\
    \text{d}^{R}_{HM}&=|R_H - R_M|
\end{split}
\end{equation}
Finally, we compute the overall aggregate distance between the model-generated story and the corresponding human-annotated story as the average of the metric-level deviations:
\begin{equation}
\label{eq:3b}
    \text{d}_{HM}=(\text{d}^{C}_{HM} + \text{d}^{G}_{HM} + \text{d}^{R}_{HM}) / 3
\end{equation}

\section{Evaluation of Existing Models}
\label{sec:4}

In this section, we evaluate and compare several state-of-the-art models using our proposed distance measure $\text{d}_{HM}$. We test the models on the popular VIST dataset, that we describe below.

\subsection{VIST dataset}
\label{sec:4_1}

VIST \citep{vist-dataset} is the first and most popular dataset for visual storytelling. The dataset includes images from Flickr albums selected by filtering titles with ``storyable'' event types (e.g., \emph{graduation ceremony}). For each of these selected albums, crowd workers constructed sequences of five images and provided corresponding five-sentence stories. The stories were then tokenized by replacing named entities including names of people, with entity types (\verb|[location]|, \verb|[organization]|) and generic placeholder tokens (\verb|[male]|, \verb|[female]|). On average, the dataset has 10.2 tokens per story and an overall vocabulary size of 18200 words. Excluding unavailable images, the dataset comprises 40071 training, 4988 validation, and 5050 test <image sequence, story> samples.

\subsection{Models}
\label{sec:4_2}

We evaluate three end-to-end models that are specifically designed and trained for the visual storytelling task. In addition, we consider two general-purpose vision-language foundation models, which are used in a zero-shot manner. 

\paragraph{GLAC Net} 
GLocal Attention Cascading Network \citep{GLACNet} is a model proposed specifically for the visual storytelling task. Adapting the standard encoder-decoder architecture, it obtains the global visual context pertaining to every image sequence position using a bi-LSTM \cite{LSTM} encoder. The obtained global context embeddings along with the individual image features (local) are, together, (GLocal) passed on to an LSTM decoder for story generation. Furthermore, to reduce redundancy in the generated sentences, GLAC Net samples multiple times from the decoder's probability distribution and selects the most frequent word from the pool.

\paragraph{AREL}
Adversarial REward Learning \cite{AREL} is another framework proposed for the visual storytelling task, which encompasses two modules: a policy model and a reward model. Similar to GLAC Net, the policy model, which uses GRUs \cite{GRU} instead of LSTMs, takes an image sequence as input and generates a story. The reward model computes a score for every input <image, sentence> pair by extracting the sentence representations using 1D-convolutional kernels and concatenating them with the corresponding pre-trained ResNet-152 \citep{resnet}) features of the images. Both modules are trained using an adversarial learning objective: the reward model is trained to discriminate between the ground truths and the generated stories; the policy model, to maximize the scores from the reward model.

\paragraph{TAPM} 
Transitional Adaptation of Pretrained Models \cite{TAPM} is a more recent approach to visual storytelling that leverages a pre-trained transformer-based language decoder. First, for every sequence position, the visual encoder pools together corresponding image features (pre-trained ResNet-101, Faster R-CNN \cite{faster-rcnn}) along with features of the images at the previous and next positions to create enriched visual context representations. Then the visual contexts are passed as input to the GPT-2\textsubscript{small} \cite{GPT2} language decoder for story generation. To bridge the semantic gap between the pre-trained image and text representations, TAPM performs an adaptation step prior to the downstream training for the visual storytelling task. Specifically, for a pre-determined number of epochs, the visual encoder parameters are fine-tuned by conditioning them on the outputs of the frozen GPT-2\textsubscript{small} decoder.

\paragraph{BLIP-2}
Unlike the previously described models, Bootstrapping Language-Image Pre-training \cite{BLIP-2} is a multimodal foundation model designed for general-purpose vision-language tasks such as visual question answering and image captioning. It connects a frozen pre-trained vision encoder and a frozen pre-trained large language model using a connector module called querying transformer (Q-Former). Q-Former contains an image and a text transformer with shared self-attention layers and learnable embeddings for querying the frozen image encoder. It is trained in two stages; in the first stage, it learns representations that align with the representations of the prompts (e.g., questions) of interest. In the second stage, its representations are fine-tuned based on the loss of the frozen language model generations (captions/answers). We adapt BLIP-2 for the visual storytelling task in a zero-shot manner by prompting the model to generate one sentence per each image in the sequence. We experiment with different settings in which the prompts contain different degrees of linguistic context.

\paragraph{LLaVA}
Similar to BLIP-2, Large Language and Vision Assistant \cite{llava-v1.6} is another general-purpose multimodal foundation model that connects large pre-trained vision encoders and large pre-trained language models. However, unlike BLIP-2, LLaVA focuses on training data and procedure as opposed to the model architecture. It is the first model that extends instruction-tuning to the language-image multimodal space. LLaVA achieves this by collecting and training on vision-language instruction-following data, constructed for <image, caption> pairs of existing datasets (e.g., COCO), by querying GPT-4 \cite{GPT4} using various in-context-learning prompts. To connect the visual features with the language embeddings, LLaVA uses a linear layer (single projection matrix) instead of Q-Former. Similar to BLIP-2, we use LLaVA to generate stories in a zero-shot manner under different linguistic context settings.

\subsection{Experimental Setup}
We generate stories for the VIST test set for all models, using greedy sampling. For GLAC Net and AREL, we leverage the publicly available model checkpoints. To obtain the model checkpoint for TAPM, we follow the proposed procedure and train the model from scratch using the VIST dataset.\footnote{Using code: \url{https://github.com/JiwanChung/tapm}.} The two general-purpose foundation models are used zero-shot, without training on the VIST dataset or the visual storytelling task. For BLIP-2, we use the version with ViT-g encoder and OPT-2.7B decoder and for LLaVA, version 1.6 with CLIP-ViT-L-336px and Mistral-7B. We prompt these models under different settings that vary with respect to the amount of linguistic and visual context given in the prompt (e.g., one image/sentence at a time vs.~all images at once). We use three prompt variations per setting and report the average of the resulting $\text{d}_{HM}$ values.\footnote{All settings and prompts are described in Appendix~\ref{appendix:s4}.}

\subsection{Results}
\label{sec:4_3}

\begin{figure}[t]
    \centering
    \includegraphics[width=\linewidth, keepaspectratio]{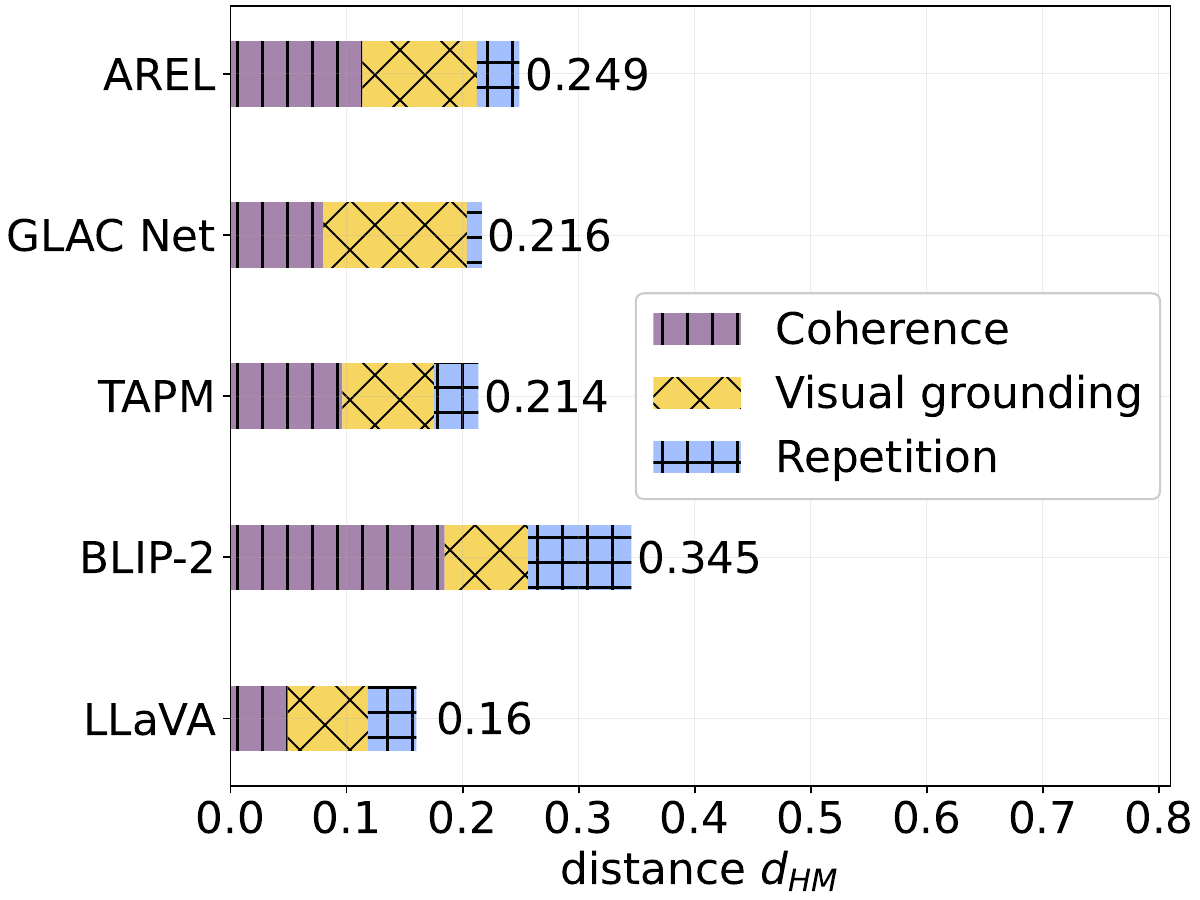}
    \caption{Distance between human- and model-generated stories in the VIST test set according to our proposed measure $\text{d}_{HM}$ (the lower the better). For BLIP-2 and LLaVA, the best setting is reported; results for all settings are provided in Fig.~\ref{fig:1sup}, Appendix~\ref{appendix:s4}.}
    \label{fig:1}
\end{figure}

Figure~\ref{fig:1} shows the distances between human-written stories and the stories generated by the models (the lower the better). Examples of model-generated stories are provided in Figure~\ref{fig:2}. In Figure~\ref{fig:1}, we observe that the stories generated by LLaVA obtain the best overall value ($\text{d}_{HM}=0.16$), followed by TAPM ($\text{d}_{HM}=0.214$). We notice that GLAC Net-generated stories exhibit the lowest distance regarding the repetition dimension. We attribute this to GLAC Net's inference phase decoding heuristic, which penalizes repetitive expressions (see Section~\ref{sec:4_2}). BLIP-2 stories are overall the farthest from stories written by humans.

Regarding the two best-performing models, LLaVA and TAPM, two points are worth highlighting. First, despite a huge difference in model size---LLaVA is a powerful 7.5B parameter foundation model, while TAPM is 50 times smaller---TAPM's $\text{d}_{HM}$ value is only slightly higher than LLaVA's. Second, LLaVA outperforms TAPM with respect to coherence and visual grounding. We hypothesize that this advantage is due to LLaVA's more powerful language and vision backbone models. In the next section, we leverage the modular architecture of TAPM to test this hypothesis.

\newlength\q
\setlength\q{\dimexpr .16\textwidth -2\tabcolsep}
\newlength\z
\setlength\z{\dimexpr .15\textwidth -2.5\tabcolsep}
\newcolumntype{L}[1]{>{\arraybackslash}p{#1}}
\begin{figure*}[t]
    \resizebox{\textwidth}{!}{\small
    \centering
    \begin{tabular}
        {@{}cccccc@{}}&\includegraphics[height=25mm]{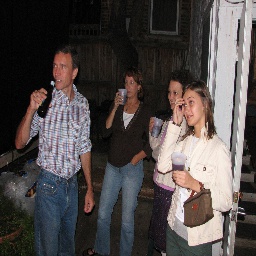}&
        \includegraphics[height=25mm]{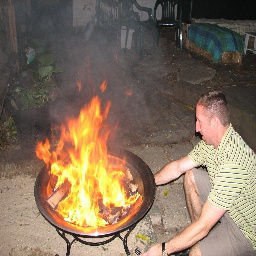}&
        \includegraphics[height=25mm]{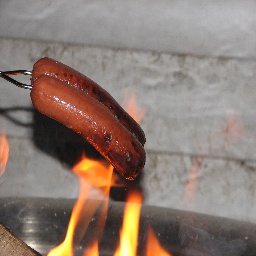}&
        \includegraphics[height=25mm]{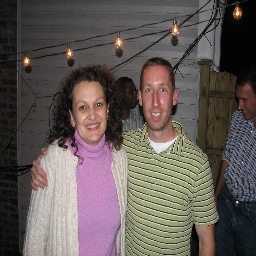}&
        \includegraphics[height=25mm]{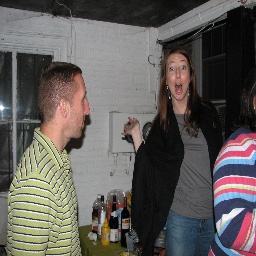}\\
        Human story & \multicolumn{5}{L{14cm}}{{we invited lots of friends for a barbeque. the fire pit was very large. we roasted hot dogs right over the flame. lots of people were happy. and there was a lot of beer too.\newline\rightline{\color{NavyBlue}C=0.993, G=0.933, R=0.968}}}\\
        \midrule
        AREL & \multicolumn{5}{L{14cm}}{the friends were having a great time at the party. the fire was <UNK> and <UNK>. the fire was <UNK> and <UNK>. the guys were having a great time. we all had a great time and had a great time.\newline\rightline{\color{NavyBlue}C=0.562, G=0.348, R=0.670}}\\\\
        GLAC Net & \multicolumn{5}{L{14cm}}{the family was having a party. they played some fire. then they had a big bonfire. everyone was happy. it was a great day.\newline\rightline{\color{NavyBlue}C=0.974, G=0.336, R=0.960}}\\\\
        TAPM & \multicolumn{5}{L{14cm}}{the group of friends got together for a bonfire. we had a lot of fun cooking. the barbecue was delicious. we took a lot of pictures. the night ended with a few drinks.\newline\rightline{\color{NavyBlue}C=0.992, G=0.597, R=0.938}}\\
        \midrule
        LLaVA & \multicolumn{5}{L{14cm}}{in the dark, a group of friends huddled around a fire, their faces lit up with the warmth of the flames. the fire crackled and roared, casting dancing shadows on their faces. one friend, a bit too eager, accidentally dropped a hot dog into the fire, causing a burst of flames and laughter. the friends cheered and clapped, their joy infectious. as the night wore on, they shared stories and laughter, the fire slowly dying down, leaving behind only the memories of their fun-filled evening.\newline\rightline{\color{NavyBlue}C=0.999, G=0.574, R=0.841}}\\\\
        BLIP-2 & \multicolumn{5}{L{14cm}}{a group of people standing outside drinking beer and talking on cell phones .  . a hot dog is being cooked on a stick over a fire .  . a group of people standing in a room with a woman making a surprised face.\newline\rightline{\color{NavyBlue}C=0.294, G=1.024, R=0.884}}\\
        \bottomrule
    \end{tabular}
    }
    \caption{An example from the VIST test set with corresponding model-generated stories and scores from individual evaluation metrics (Coherence: \color{NavyBlue}C\color{black}, Visual grounding: \color{NavyBlue}G\color{black}, Repetition: \color{NavyBlue}R\color{black}).}
    \label{fig:2}
\end{figure*}

\section{Model Analysis and Improvements}
\label{sec:5}

In Section~\ref{sec:4}, we showed that the $\text{d}_{HM}$ obtained by the visual storytelling-specific TAPM model is only slightly higher than that of LLaVA, a 50-times larger foundation model. In particular, we notice that LLaVA has an advantage over TAPM in two dimensions, i.e., visual grounding and coherence. We hypothesize that this advantage is due to the model's better language and vision backbone models---LLaVA builds on a pre-trained, transformer-based language model and image processor. Thus, we leverage the modular architecture of TAPM and test whether we can obtain better results (lower distances) by replacing its original language and vision components with models similar to those embedded in LLaVA, while keeping the number of parameters significantly lower. To test whether the results we obtain are consistent across datasets, we perform this analysis on both VIST and VWP~\cite{VWP}.

\subsection{Updating the Language Component}

By default, TAPM uses GPT-2\textsubscript{small} as its story decoder. Here, we replace this language model with \textsc{Llama 2}~\cite{llama2}, an auto-regressive (decoder-only) large language model pre-trained via maximum likelihood objective for next-token prediction on massive amounts of publicly available online data. Thanks to the large context window it can take as input and the Grouped Query Attention mechanism to speed up processing during decoding~\cite{GQA}, this LM is currently the state-of-art on various downstream tasks in the MMLU benchmark~\cite{MMLU}. Here, we use the 4-bit quantized~\cite{QLoRA} pre-trained version of the \textsc{Llama 2 7B} model and adapt the parameter dimensions of TAPM's visual encoder component to match the updated language decoder. To ensure computational and memory efficiency, we employ the LoRA~\cite[Low-rank Adapter;][]{LoRA} fine-tuning approach (that allows for updating only a subset of the model's parameters) and only target the multi-head self-attention blocks---$W^Q, W^K, W^V, W^O$~\cite{transformer}---of the language model during training.\footnote{Further details on training are provided in Appendix~\ref{appendix:s5}.} Henceforth, we refer to this upgraded TAPM model as (+\textsc{Llama 2}).\looseness-1

\subsection{Updating the Vision Component}

The original TAPM model uses pre-trained ResNet-101 and Faster R-CNN for extracting the image-level and object-level features, respectively. We supplement the image-level ResNet features by concatenating them with representations extracted using pre-trained Vision Transformer model~\cite[ViT\textsubscript{base};][]{ViT}. ViT\textsubscript{base} leverages the transformer architecture for image processing and is pre-trained on the ImageNet-21K \cite{ImageNet21K} data. Features extracted using ViT\textsubscript{base} have been shown to improve the performance of models on several computer vision tasks. Henceforth, we refer to this upgraded TAPM model as (+ViT).

\subsection{Experimental Setup}
We train the (+ViT) and (+\textsc{Llama 2}) models from scratch using the VIST data training and validation splits for 15 epochs, and obtain the results on the test split. We conduct the same experiment for the VWP dataset independently, by following the procedure proposed by the authors of the dataset. As mentioned in Section~\ref{sec:2_1}, VWP includes sequences of 5-10 images constructed using frames obtained from MovieNet~\cite{MovieNet}. Following the same procedure used by the authors, we split the dataset into 9606 training, 849 validation, and 586 test <image sequence, story> samples and preprocess the text to replace recognized named entities with entity types and placeholders. To compare the performance of the improved TAPM models against LLaVA, we use LLaVA to generate stories for the VWP test set by prompting it under the visual context setting.

\subsection{Results}

\begin{figure}[t!]
    \centering
    \includegraphics[width=\linewidth, keepaspectratio]{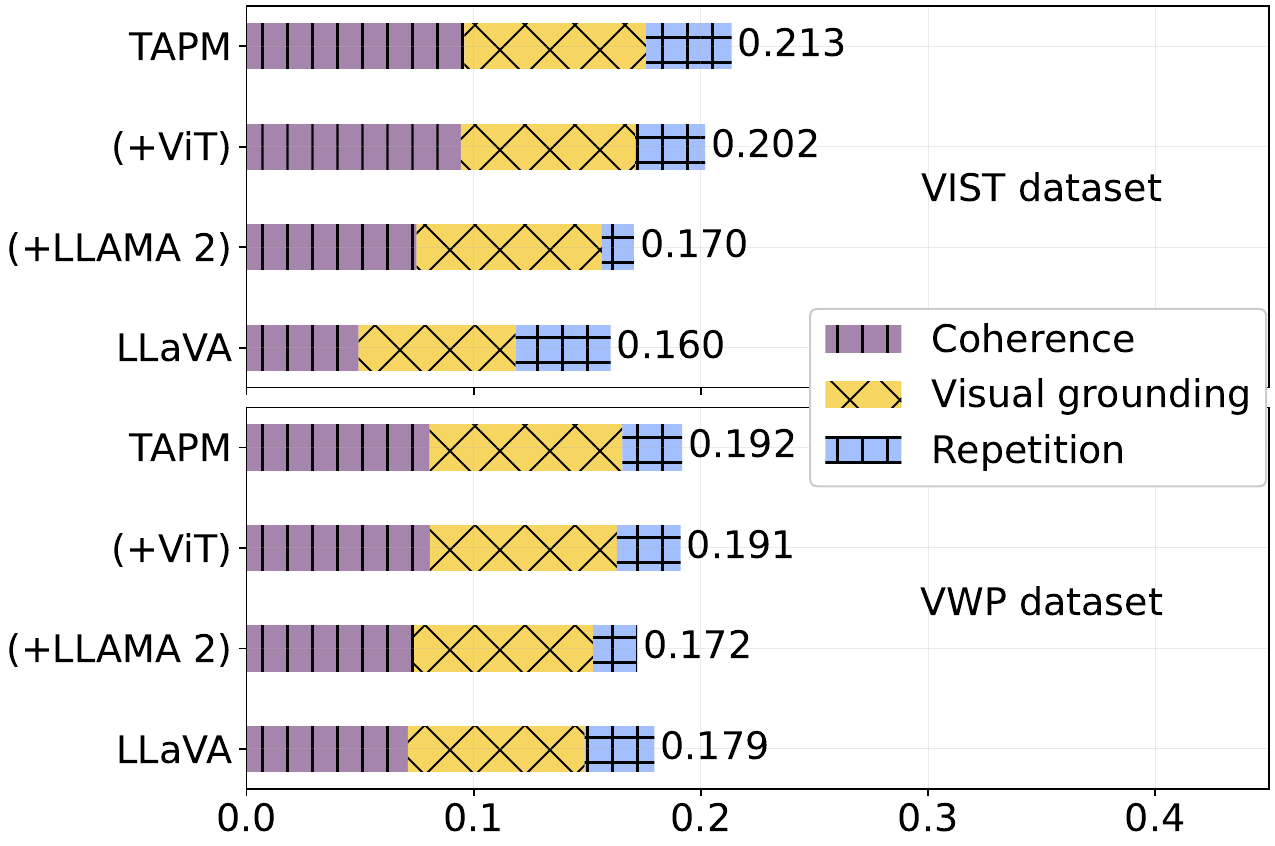}
    \caption{For the TAPM model and its upgraded versions (+\textsc{Llama 2}) and (+ViT), we compute the $\text{d}_{HM}$ values on two datasets---VIST and VWP.}
    \label{fig:3}
\end{figure}

\begin{figure}[ht]
    \centering
    \includegraphics[width=\linewidth, keepaspectratio]{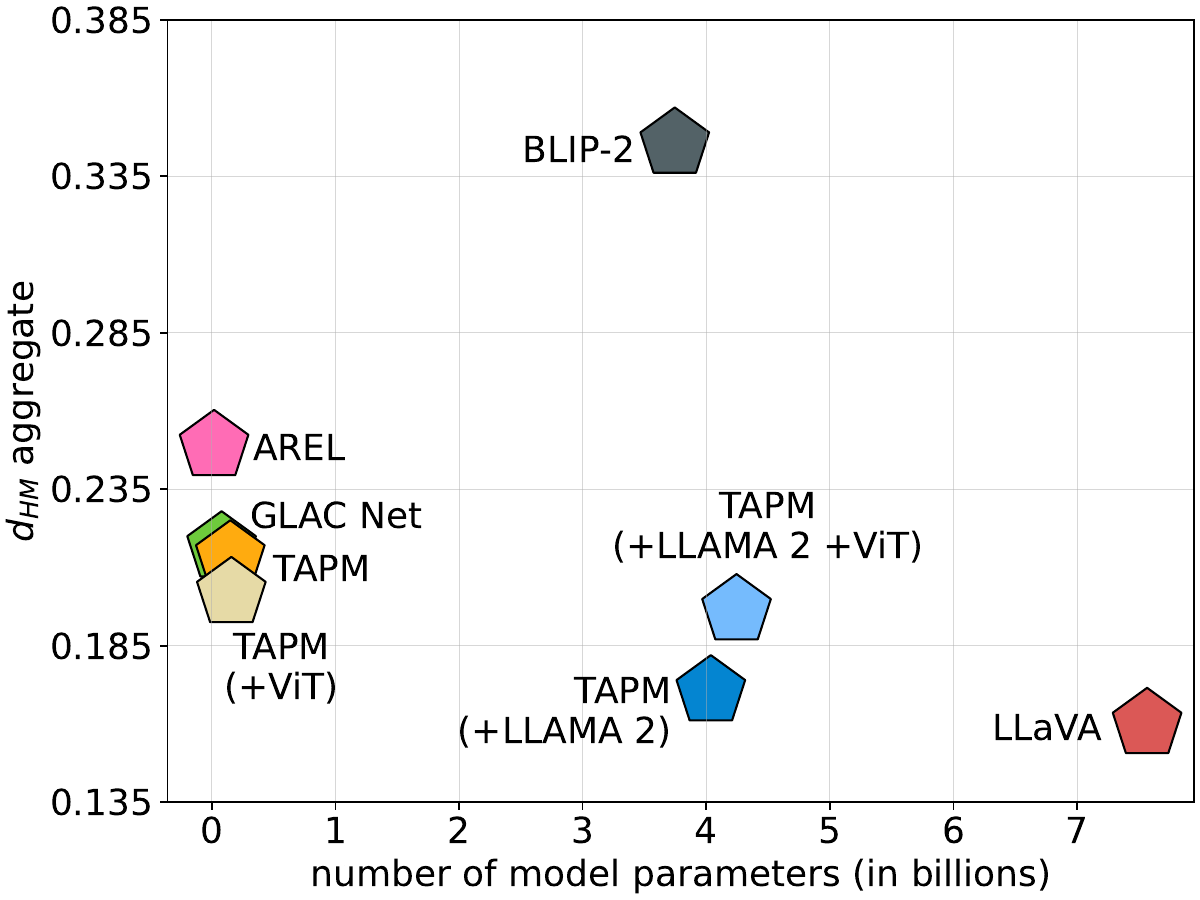}
    \caption{Comparison between the number of model parameters and their corresponding $\text{d}_{HM}$ values on the VIST test set (the lower the better). We observe that as the models increase in size, the scores for the stories they generated get closer to scores of human annotations.}
    \label{fig:4}
\end{figure}

Figure~\ref{fig:3} shows the $\text{d}_{HM}$ values for the LLaVA model and for all the different versions of the TAPM model on both the VIST and the VWP datasets. Firstly, we observe that compared to the original TAPM model, (+\textsc{Llama 2}) generates stories that are closest to human stories in terms of both coherence and repetition, whereas stories by the (+ViT) version are closest along the dimension of visual grounding. These results are in line with our intuitions regarding the influence that both modalities have on different aspects of visual story evaluation. For this analysis, we also considered a version of the TAPM model in which both the language and vision components are jointly updated (+\textsc{Llama 2}, +ViT).
However, (+\textsc{Llama 2}, +ViT) consistently under-performed on our distance measure $\text{d}_{HM}$ compared to the (+\textsc{Llama 2}) and (+ViT) versions, and was only marginally better than the original TAPM model. Despite the significant difference in the number of parameters, we notice that (+\textsc{Llama 2}) achieves performance on par with LLaVA (see Figure~\ref{fig:4}).\looseness-1

We observe similar results for the TAPM models on the VWP dataset---(+ViT) obtains the lowest distance in terms of visual grounding and (+\textsc{Llama 2}) obtains the lowest distance in terms of both coherence and repetition. Furthermore, on the VWP dataset, the (+\textsc{Llama 2}) model achieves the overall lowest $\text{d}_{HM}$, performing better than the LLaVA model. These results quantitatively indicate that the stories generated by the models are close to the corresponding human-written stories. To better understand if the metrics are capable of effectively comparing stories generated by models with human-written ones, we conduct a qualitative evaluation, that we describe in the next section.\looseness-1

\section{Qualitative Analysis and Discussion}
\label{sec:6}

Our results suggest that the stories generated by the best-performing models according to $\text{d}_{HM}$ are very close to human levels of visual grounding, coherence, and degree of repetition. To test whether this aligns with the perceived overall quality of the stories, we conduct a qualitative human evaluation. We consider the VIST stories generated by the two models---TAPM (+\textsc{Llama 2}) and LLaVA---that achieve the best performance in terms of $\text{d}_{HM}$ (lowest distances). For each model, we select 100 generated stories, that we randomly sample using the distribution of $\text{d}_{HM}$ values on the VIST test set.\footnote{See Figure~\ref{fig:5supA} in Appendix~\ref{appendix:s6}.} We then provide annotators with <image sequence, model-story> pairs along with corresponding human-written stories, and ask them to assess whether one is better than the other, or whether both are similarly fine or bad. Five annotators unrelated to the project participated in the task.\footnote{Instructions and additional details in Appendix~\ref{appendix:s6}.}

Figure~\ref{fig:5} shows the judgments obtained from the annotators for the two models. We observe that humans consistently prefer human-written stories over stories generated by the models. This happens slightly more often for TAPM- than for LLaVA-generated stories (46\% vs~57\% of cases). Similarly, LLaVA-generated stories are more frequently preferred over human stories than stories generated by TAPM, although this happens very seldom for both models (16.6\% vs.~6.2\%). These patterns are in line with the results obtained with our distance metric, according to which LLaVA has a slight advantage over TAPM. Indeed, we observe that the average $\text{d}_{HM}$ is higher when the human story is judged as better than when both stories are perceived as having similar quality; this difference is statistically significant for TAPM (avg $\text{d}_{HM}$=0.279 vs.~0.243; $t=2.19, p < 0.03$) and not so for LLaVA  (avg $\text{d}_{HM}$=0.240 vs.~0.226; $t=1.23, p> 0.05$). Overall, this suggests that our quantitative evaluation effectively captures key aspects of visual stories.

\begin{figure}[t!]
    \centering
    \includegraphics[width=\columnwidth, keepaspectratio]{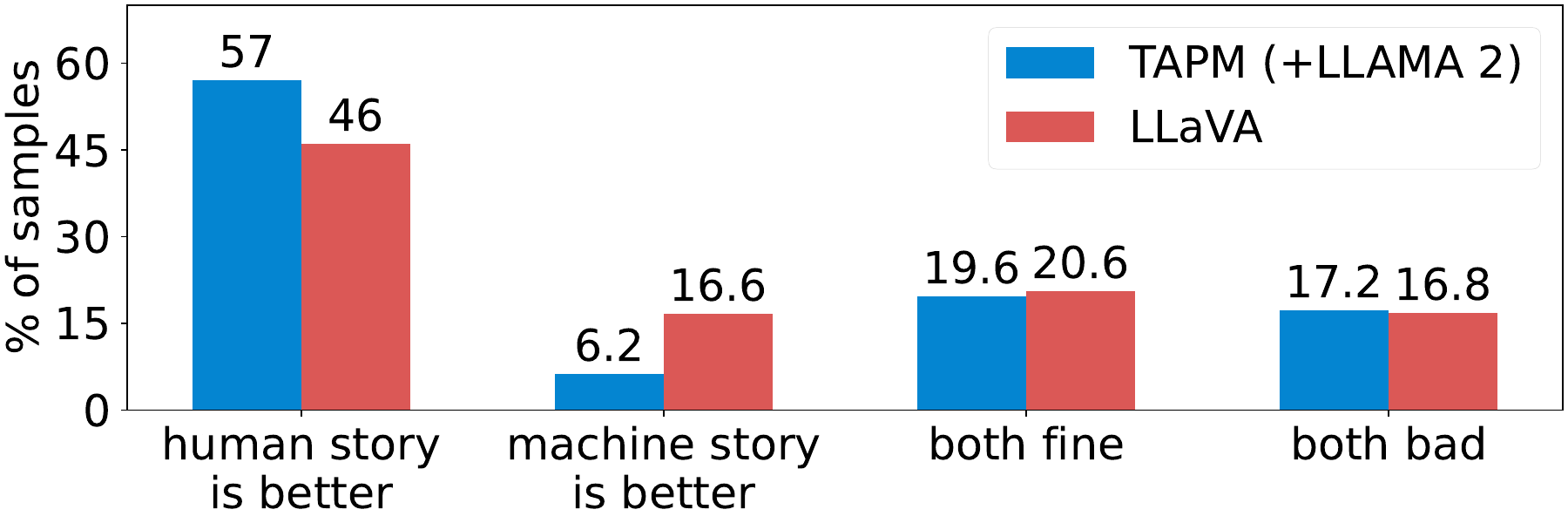}
    \caption{Aggregated judgments (5 human evaluators per model) comparing human stories in the VIST test set to stories generated by the two best-performing models according to our distance measure $\text{d}_{HM}$.}
    \label{fig:5}
\end{figure}

Yet, preference for human-written stories suggests that a good story involves more than just visual grounding, coherence, and limited repetition. To get more insight into this issue, we asked the annotators in the study to describe the properties they considered when comparing stories. Some annotators reported that they employed their subjective world knowledge in differentiating between creative and hallucinated stories. Also, stories that describe events without an overarching narrative were judged as bad despite being locally coherent, well-grounded, and non-repetitive. In addition, annotators generally preferred stories that contained phrases or sentences expressing emotions. We provide examples of these cases in Appendix~\ref{appendix:s6}, which illustrate the limitations of current evaluation metrics in fully capturing hallucinations, relevant emotions, and creative expressions in the generated stories. Therefore, there is scope for a lot of further work in this evaluation domain.

\section{Conclusion}

We proposed a novel human-centric method ($\text{d}_{HM}$) for evaluating the quality of model-generated stories in terms of their closeness to human-written stories along coherence, visual grounding, and non-redundancy. Using our proposed method, we compared various models and found that the large foundation model LLaVA obtains slightly better results than the (50 times smaller) best visual storytelling model TAPM. We showed that upgrading the components of TAPM boosts its performance (lowers its $\text{d}_{HM}$) across multiple datasets, confirming the advantage of leveraging last-generation language and vision pre-trained models. Finally, we conducted a qualitative human evaluation to explore whether these quantitative findings align with the overall perceived story quality. We observed that human judgments align with our quantitative evaluation. Yet, humans still prefer human-written stories over model-generated ones, which suggests that capturing coherence, visual grounding, and non-repetitiveness may not (yet) be the whole story.

\section*{Limitations}

The human evaluation analysis we performed is arguably small-scale. As such, we cannot rule out that carrying it out with more annotators and a larger set of stimuli may lead to different patterns of results. Second, the number of models we experimented with is quite limited, which is an obvious limitation of this work. Yet, we defend the selection we made, aimed at a trade-off between the limited availability of computational resources and the inclusion of various architecture families. Finally, we experimented with only two visual storytelling datasets, both in English and both Western-centric. This is a limitation of our work, which is, however, due to the unavailability of other datasets with more diverse language and cultural backgrounds. We strongly support the creation of such resources.

\section*{Acknowledgments}

We are immensely grateful to the participants of the qualitative evaluation study and to our colleagues at the Dialogue Modelling Group for their invaluable inputs at different stages of this work. AKS was supported by the TIMELY project under the EU-H2020 grant 101017424. RF was supported by the European Research Council (ERC Consolidator Grant DREAM 819455).

\bibliography{anthology,custom}
\newpage
\appendix

\section{Zero-Shot Settings and Prompts}
\label{appendix:s4}

We generated stories using the two foundation models---BLIP-2, LLaVA---zero-shot by prompting them under different settings. In the \emph{visual} context setting, models received the entire image sequence (with its images horizontally concatenated) as input, along with a prompt designed for this setting. In the \emph{linguistic} context settings, models generated one sentence per image in the sequence. We obtained the results of contextualizing the prompt with different degrees of linguistic information---sentence generated for the previous image in the sequence (\emph{prev sentence}), concatenated prefix of sentences generated for all the preceding images in the sequence (\emph{all sentences}). We used three variations of the prompts and reported the average of the resulting $\text{d}_{HM}$ values. Prompts used in all the settings are as provided below:
\vspace*{0.1cm}
\begin{lstlisting}[language=python]
# prompts used for visual context

P1 = '[INST] <image>\nWrite a story using exactly five sentences for this image sequence. Do not use more than five sentences. [/INST]'
P2 = '[INST] <image>\nGenerate a story consisting of five sentences for this image sequence. Use only five sentences and not more. [/INST]'
P3 = '[INST] <image>\nOutput a story about this sequence of images using only five sentences. Make sure the story does not include more than five sentences. [/INST]'
\end{lstlisting}
\begin{lstlisting}[language=python]
# prompts used for linguistic contexts:
## {prev sentence} or {all sentences}

P1 = '[INST] <image>\nUsing this image, add one sentence to the following story: {context}<s> [/INST]'
P2 = '[INST] <image>\nGiven this image, write one sentence as an addition to the following story: {context}<s> [/INST]'
P3 = '[INST] <image>\nAdd one sentence based on this image to the following story: {context}<s> [/INST]'
\end{lstlisting}

\begin{figure}[t!]
    \centering
    \includegraphics[width=\linewidth, keepaspectratio]{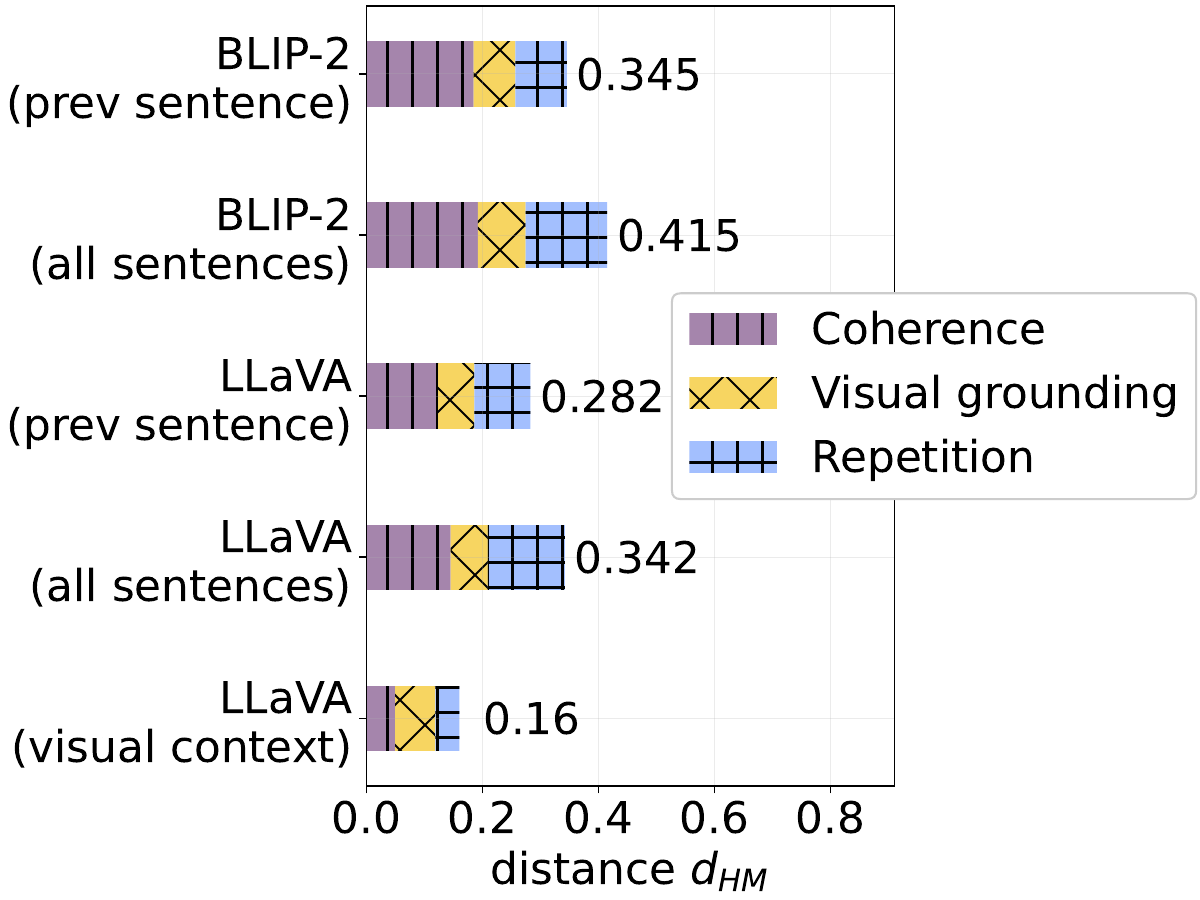}
    \caption{$\text{d}_{HM}$ values (the lower the better) for the stories generated by the BLIP-2 and LLaVA models under different zero-shot settings.}
    \label{fig:1sup}
\end{figure}

Figure~\ref{fig:1sup} shows the $\text{d}_{HM}$ values for both the models on the VIST test set, across all zero-shot settings. Besides the LLaVA model under the \emph{visual} context setting, models obtained significantly high $\text{d}_{HM}$ values in other settings. We observed that under the \emph{visual} context setting, the BLIP-2 model failed to generate any stories.

\section{Improvements to TAPM}
\label{appendix:s5}

\begin{figure*}[t]
    \centering
    \subcaptionbox{$\text{d}_{HM}$ aggregate}{\includegraphics[width=0.45\textwidth]{images/dHM_vs_msize.pdf}}
    \hfill
    \subcaptionbox{Coherence ($\text{d}^C_{HM}$)}{\includegraphics[width=0.45\textwidth]{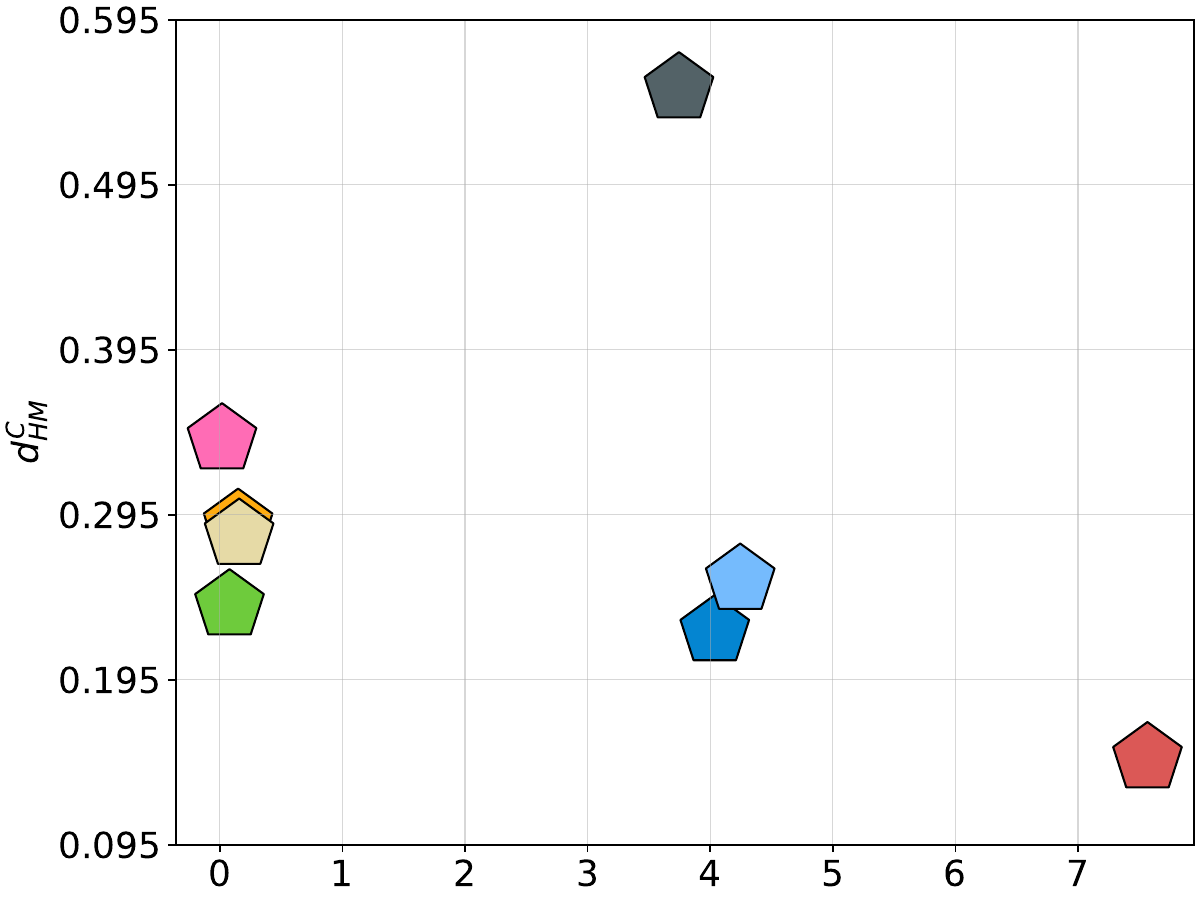}}
    \subcaptionbox{Visual grounding ($\text{d}^G_{HM}$)}{\includegraphics[width=0.45\textwidth]{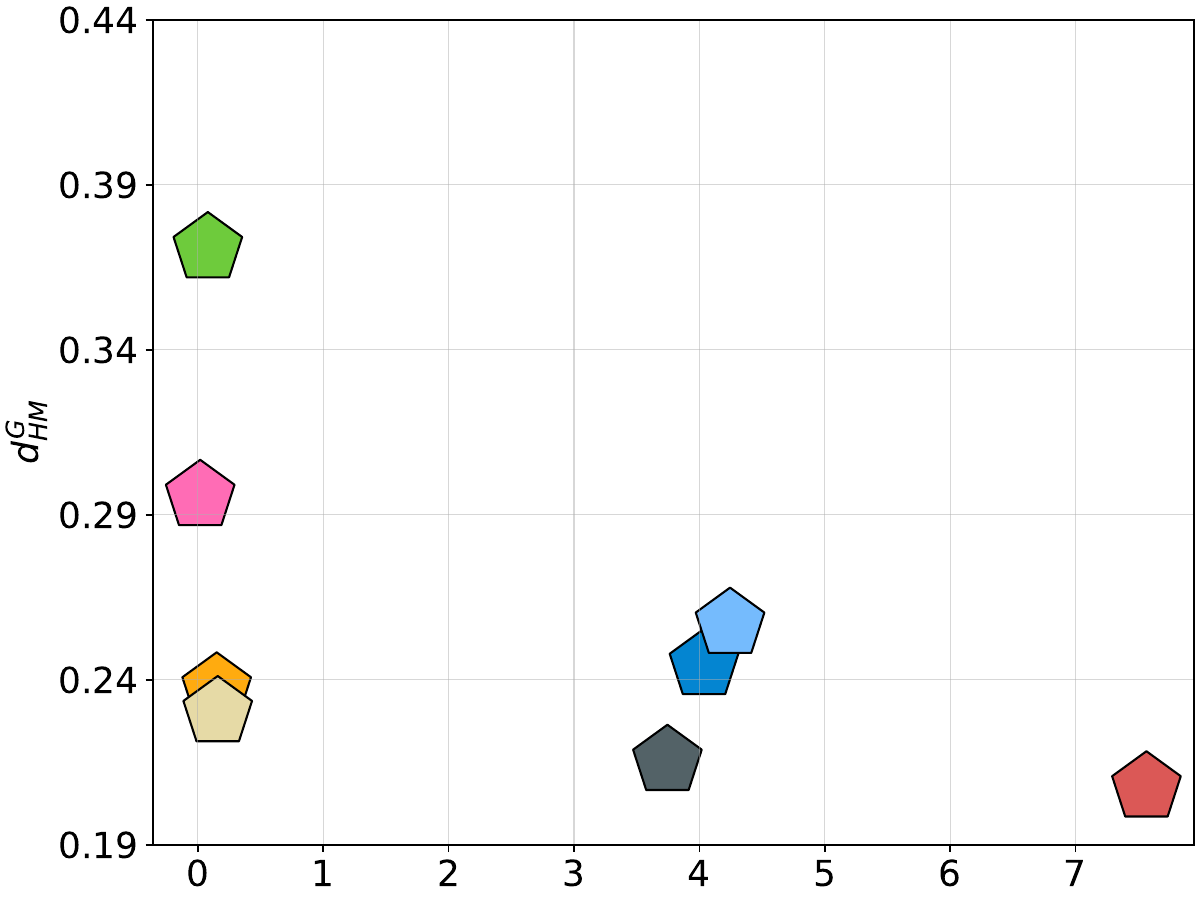}}
    \hfill
    \subcaptionbox{Repetition ($\text{d}^R_{HM}$)}{\includegraphics[width=0.45\textwidth]{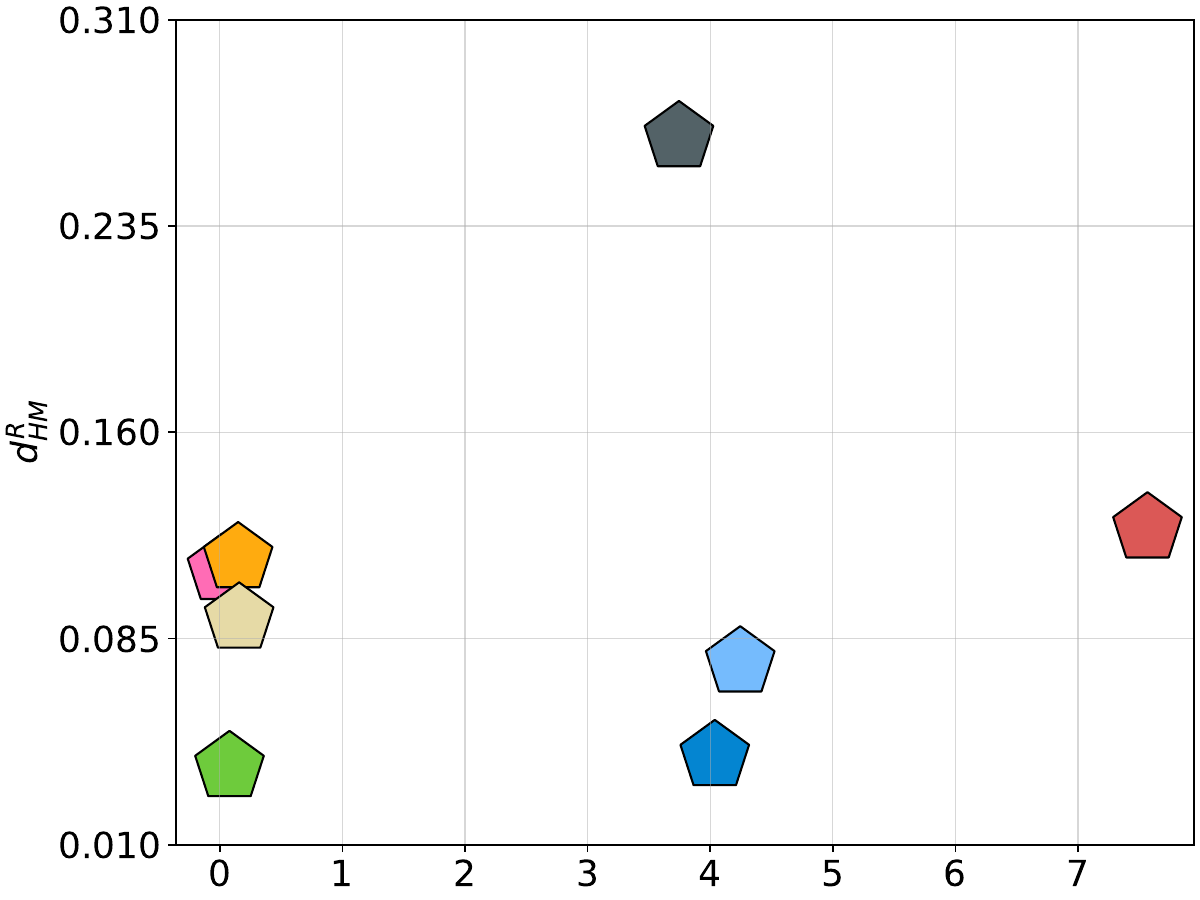}}
    \caption{
    Comparison between the number of model parameters and their corresponding $\text{d}_{HM}$ values on the VIST test set (the lower the better). Plot \textbf{(a)} compares the overall aggregate distance and plots \textbf{(b)} through \textbf{(d)} show individual metric-level distances. We note that the stories generated by the overall best performing model---LLaVA---obtains the closest distance to human stories in terms of coherence and visual grounding.}
    \label{fig:4sup}
\end{figure*}

To train the (+\textsc{Llama 2}) version, we adapted the model's dimensionality from 768 to 4096. We used a batch size of 2 and trained the model for 18 epochs---3 for the pre-task adaptation and 15 for the downstream task training. We used the 4-bit quantized pre-trained version of (+\textsc{Llama 2}) 7B and employed LoRA fine-tuning. We configured the fine-tuning with a scaling parameter of $\alpha=8$ and a rank of $r=8$, based on the evidence provided in \citet{LoRA}. To generate stories using the trained model, we used greedy decoding. For completing 1 epoch of training and inference on the VIST dataset, the (+\textsc{Llama 2}) model approximately used 8 compute hours of 1 Nvidia A100 (40GB) GPU. For all the models we used in Sections~\ref{sec:4} and~\ref{sec:5}, Figure~\ref{fig:4sup} compares the relationship between the size of the models and the corresponding distances they obtain along the aspects of coherence, visual grounding, and repetition.\looseness-1

\section{Human Evaluation}
\label{appendix:s6}

\begin{figure*}
    \fbox{\centering\includegraphics[scale=0.58, keepaspectratio, width=\linewidth]{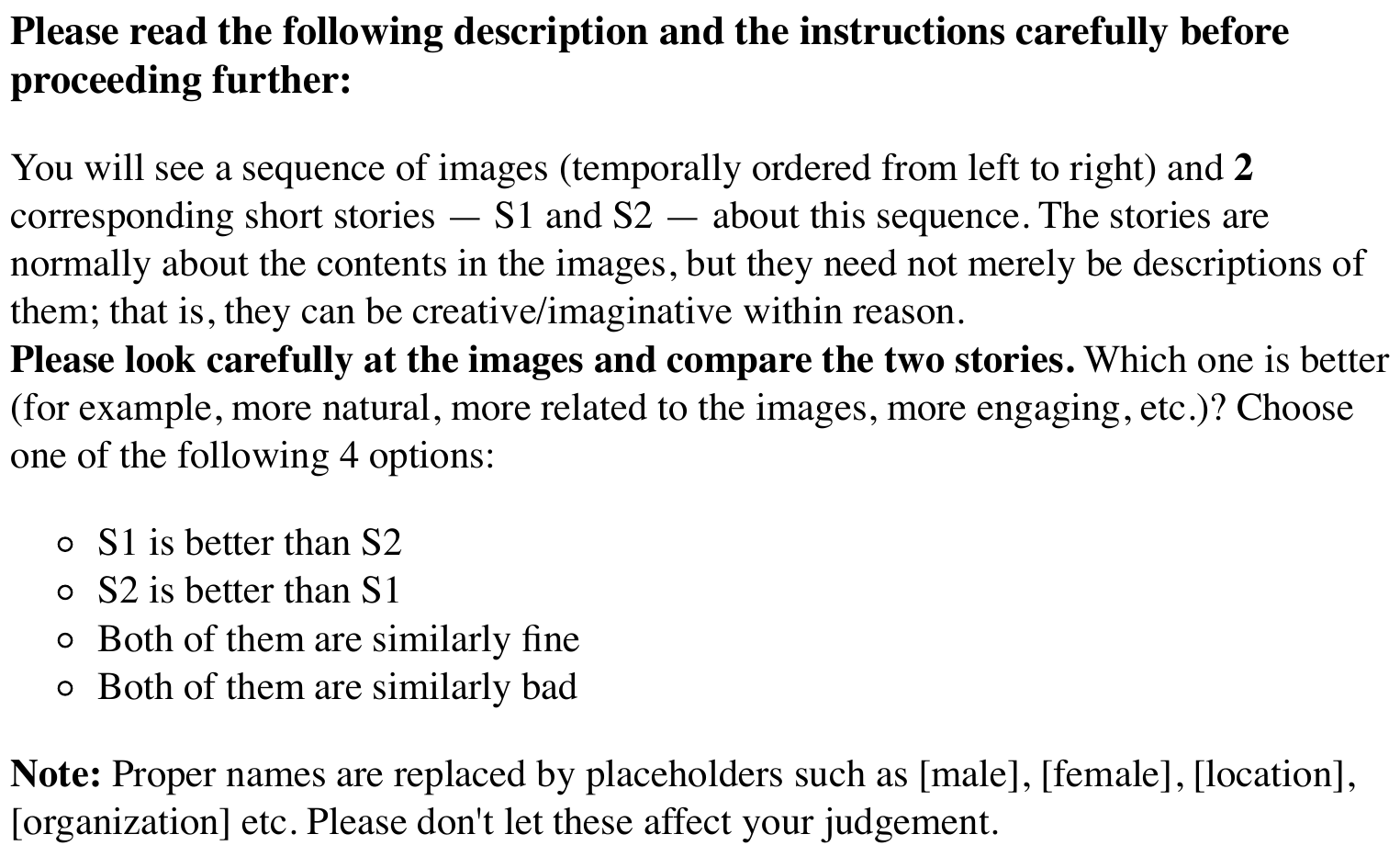}}
    \caption{Human evaluation task description and instructions.}
    \label{fig:inst}
\end{figure*}

\begin{figure*}[th]
    \resizebox{\textwidth}{!}{\small
    \centering
    \begin{tabular}
        {@{}cccccc@{}}&\includegraphics[height=25mm]{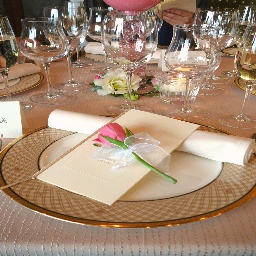}&
        \includegraphics[height=25mm]{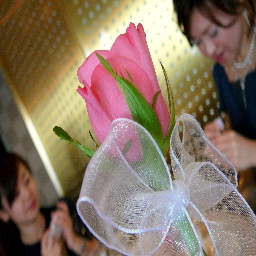}&
        \includegraphics[height=25mm]{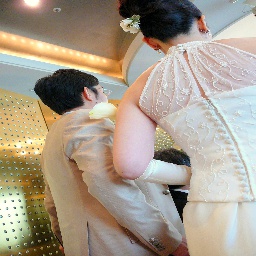}&
        \includegraphics[height=25mm]{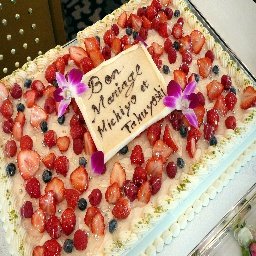}&
        \includegraphics[height=25mm]{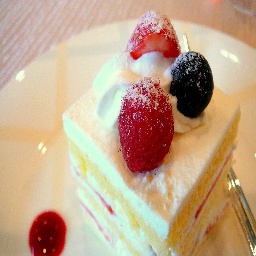}\\
        \\
        \textbf{\color{red}{S1}} & \multicolumn{5}{L{14cm}}{the table is set for a dinner party. the guests were amazed by the decorations. the food was delicious. they are making a wish. then they had dessert.\newline\rightline{\color{NavyBlue}C=0.956, G=1.341, R=0.938}}\\\\[+0pt]
        \textbf{\color{green}{S2}} & \multicolumn{5}{L{14cm}}{our first meal together as a married couple was as beautiful as it was delicious. the roses they sent over were of the highest quality like the ingredients in their food. the personal touches like the unique cake designs made it even more special. we were on our way to married life in high class style. and this was literally the icing on the cake.\newline\rightline{\color{NavyBlue}C=0.504, G=0.815, R=0.901}}\\\\[+0pt]
    \end{tabular}
    }
    \caption{Example depicting emotions: all annotators evaluated story \textbf{\color{green}{S2}}\color{black}~as better than story \textbf{\color{red}{S1}}. \color{NavyBlue}C\color{black}, \color{NavyBlue}G\color{black}, and \color{NavyBlue}R \color{black} denote the coherence, visual grounding, and repetition scores respectively.}
    \label{fig:A}
\end{figure*}

\begin{figure*}[htb]
    \resizebox{\textwidth}{!}{\small
    \centering
    \begin{tabular}
        {@{}cccccc@{}}&\includegraphics[height=25mm]{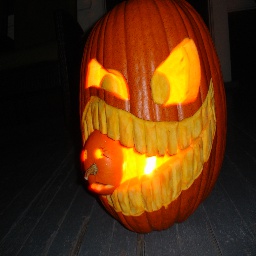}&
        \includegraphics[height=25mm]{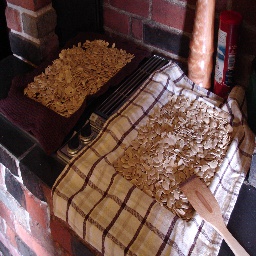}&
        \includegraphics[height=25mm]{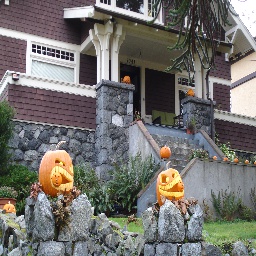}&
        \includegraphics[height=25mm]{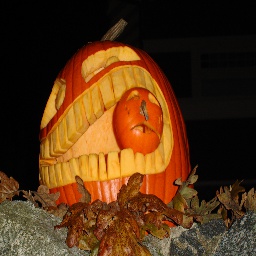}&
        \includegraphics[height=25mm]{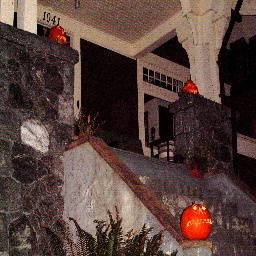}\\
        \\
        \textbf{\color{green}{S1}} & \multicolumn{5}{L{14cm}}{it's halloween and the pumpkins are being carved. i bought a lot of food for it. the house has a lot of decorations. the pumpkin was carved with a scary face. the pumpkins are lit up inside.\newline\rightline{\color{NavyBlue}C=0.980, G=1.661, R=0.942}}\\\\[-0pt]
        \textbf{\color{red}{S2}} & \multicolumn{5}{L{14cm}}{the pumpkin was angry. someone had stolen all of his seeds. he waited patiently in front of the house for night to fall. once it was night time he made his move. he proceeded into the house to finally get his revenge. there were no survivors.\newline\rightline{\color{NavyBlue}C=0.647, G=0.703, R=0.971}}\\\\[-0pt]
    \end{tabular}
    }
    \caption{Example with hallucinations: 4 annotators (of 5) either selected story \textbf{\color{green}S1}\color{black}~as better or evaluated story \textbf{\color{red}S2}\color{black}~as bad. \color{NavyBlue}C\color{black}, \color{NavyBlue}G\color{black}, and \color{NavyBlue}R \color{black} denote the coherence, visual grounding, and repetition scores respectively.}
    \label{fig:B}
\end{figure*}

\begin{figure*}[b]
    \resizebox{\textwidth}{!}{\small
    \centering
    \begin{tabular}
        {@{}cccccc@{}}&\includegraphics[height=25mm]{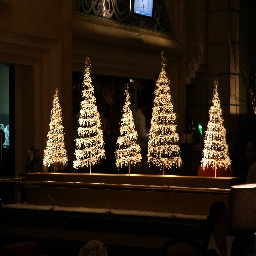}&
        \includegraphics[height=25mm]{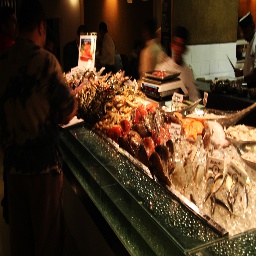}&
        \includegraphics[height=25mm]{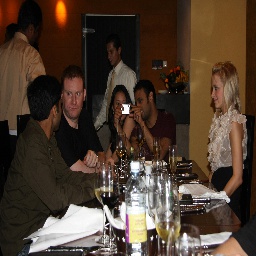}&
        \includegraphics[height=25mm]{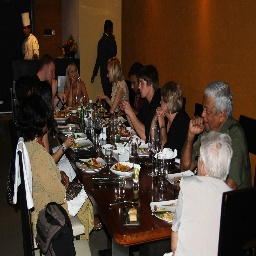}&
        \includegraphics[height=25mm]{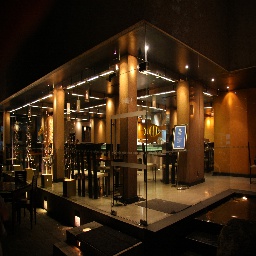}\\
        \\
        \textbf{\color{green}{S1}} & \multicolumn{5}{L{14cm}}{our business hosted its winter holiday party at a hotel. the food was served as a buffet. all the young workers sat at one end of the table. the older co-workers sat at the other end of the table. in the end, both young and old co-workers stayed until closing time for the hotel restaurant.\newline\rightline{\color{NavyBlue}C=0.999, G=0.827, R=0.881}}\\\\[-0pt]
        \textbf{\color{red}{S2}} & \multicolumn{5}{L{14cm}}{in the first image, a man stands in awe of a beautifully decorated Christmas tree. the second image captures a lively dinner party, with guests seated around a table laden with food and wine. the third image shows a bustling restaurant kitchen, where chefs are busy preparing meals. the fourth image reveals a cozy dining room, where a family enjoys a quiet meal together. the final image presents a grand banquet hall, filled with guests and adorned with elegant decorations.\newline\rightline{\color{NavyBlue}C=0.996, G=1.576, R=0.867}}\\[-0pt]
    \end{tabular}
    }
    \caption{Example with an overarching narrative: all annotators evaluated story \textbf{\color{green}{S1}}\color{black}~as better than story \textbf{\color{red}{S2}}. \color{NavyBlue}C\color{black}, \color{NavyBlue}G\color{black}, and \color{NavyBlue}R \color{black} denote the coherence, visual grounding, and repetition scores respectively.}
    \label{fig:C}
\end{figure*}

Five annotators unrelated to the project participated in the study voluntarily. We obtained their consent about using the data collected as part of the study for academic research. Upon expressing their consent, the annotators received access to an interface with the task description and instructions, shown in Figure~\ref{fig:inst}. Along with the instructions, we provided one example for each of the four possible options in the task. After completing the task, we asked each annotator to describe the properties they considered for comparing the stories. Specifically, we asked the following question: ``\emph{What are the properties of a story that made you select it as being better?}''. Based on the responses and judgements obtained from the annotators (see Section~\ref{sec:6}), we report qualitative examples in Figures~\ref{fig:A},~\ref{fig:B}, and~\ref{fig:C}.


\begin{figure*}[b]
    \centering
    \subcaptionbox{TAPM (+\textsc{Llama 2})}{\includegraphics[width=0.43\textwidth]{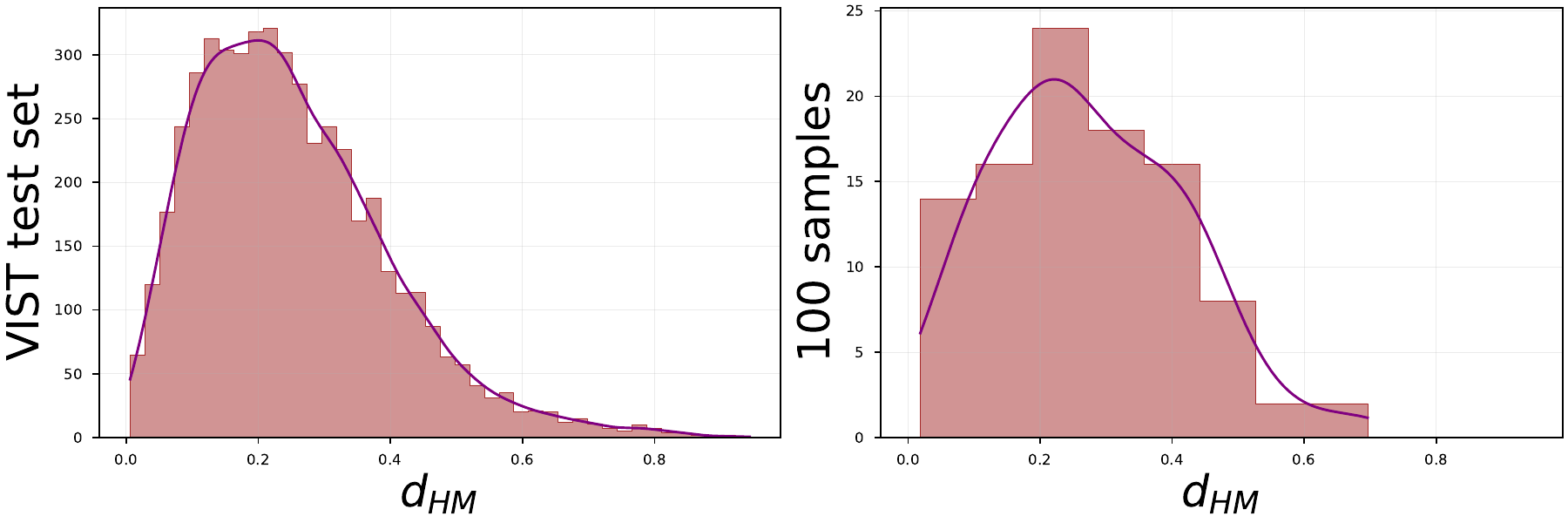}}
    \hfill
    \subcaptionbox{LLaVA}
    {\includegraphics[width=0.43\textwidth]{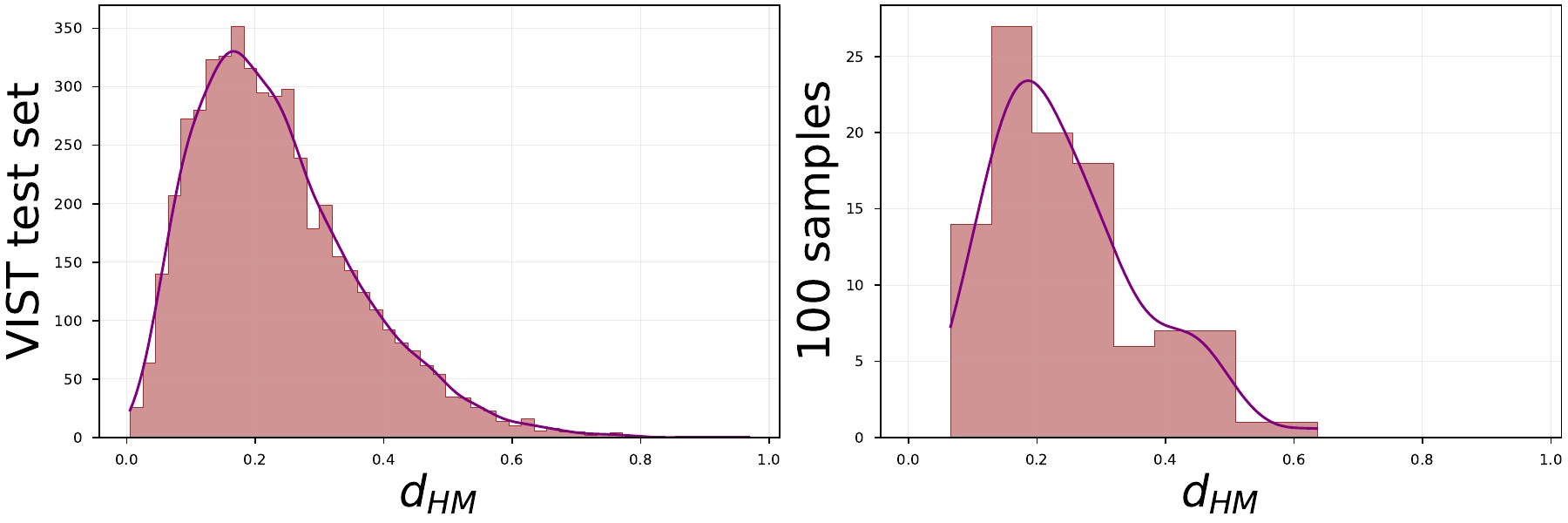}}
    \caption{$\text{d}_{HM}$ distributions for the VIST test set (left) and for the 100 randomly sampled instances (right).}
    \label{fig:5supA}
\end{figure*}

\end{document}